\documentclass[10pt,twocolumn,letterpaper]{article}

\usepackage{cvpr}              

\usepackage[dvipsnames]{xcolor}

\definecolor{cvprblue}{rgb}{0.21,0.49,0.74}
\usepackage[pagebackref,breaklinks,colorlinks,citecolor=cvprblue]{hyperref}

\def\paperID{20} 
\def\confName{CVPR}
\def\confYear{2024}

\usepackage{times}
\usepackage{epsfig}
\usepackage{graphicx}
\usepackage{amsmath}
\usepackage{amssymb}

\usepackage{dsfont}  
\usepackage{booktabs}

\usepackage{colortbl}
\usepackage{xcolor}         
\definecolor{Gray}{gray}{0.9}

\newcommand{\jie}[1]{#1}

\definecolor{myGreen}{rgb}{0, .6, .0}
\hypersetup{colorlinks,breaklinks,anchorcolor=darkblue,citecolor=myGreen}

\usepackage[capitalize]{cleveref}
\crefname{section}{Sec.}{Secs.}
\Crefname{section}{Section}{Sections}
\Crefname{table}{Table}{Tables}
\crefname{table}{Tab.}{Tabs.}

\begin{document}

\title{What’s in a Name? Beyond Class Indices for Image Recognition}

\author{Kai Han$^{1}$\footnotemark[1]
\quad Xiaohu Huang$^{1}$\footnotemark[1]
\quad Yandong Li$^{2}$\footnotemark[1]
\quad Sagar Vaze$^{3}$\footnotemark[1]
\quad Jie Li$^1$
\quad Xuhui Jia$^{2}$\\
$^1$ The University of Hong Kong
\quad $^2$ Google Research
\quad $^3$ VGG, University of Oxford\\
{\tt\small kaihanx@hku.hk \quad hxh247@connect.hku.hk \quad yandongli@google.com } \\ {\tt\small \quad sagar@robots.ox.ac.uk \quad jieli23@hku.hk \quad xhjia@google.com}
}

\maketitle

\renewcommand{\thefootnote}{\fnsymbol{footnote}}
\footnotetext[1]{Equal contribution.}

\begin{abstract}
    Existing machine learning models demonstrate excellent performance in image object recognition after training on a large-scale dataset under full supervision. However, these models only learn to map an image to a predefined class index, without revealing the actual semantic meaning of the object in the image.
    In contrast, vision-language models like CLIP are able to assign semantic class names to unseen objects in a `zero-shot' manner, though they are once again provided a pre-defined set of candidate names at test-time. 
    In this paper, we reconsider the recognition problem and task a vision-language model with assigning class names to images given only a large (essentially unconstrained) vocabulary of categories as prior information.
    We leverage non-parametric methods to establish meaningful relationships between images, allowing the model to automatically narrow down the pool of candidate names. Our proposed approach entails iteratively clustering the data and employing a voting mechanism to determine the most suitable class names.
    Additionally, we investigate the potential of incorporating additional textual features to enhance clustering performance. To achieve this, we employ the CLIP vision and text encoders to retrieve relevant texts from an external database, which can provide supplementary semantic information to inform the clustering process.
    Furthermore, we tackle this problem both in unsupervised and partially supervised settings, as well as with a coarse-grained and fine-grained search space as the unconstrained dictionary. 
    Remarkably, our method leads to a roughly 50\% improvement over the baseline on ImageNet in the unsupervised setting. 
\end{abstract}

\begin{figure*}
    \centering
    \includegraphics[width=0.9\linewidth]{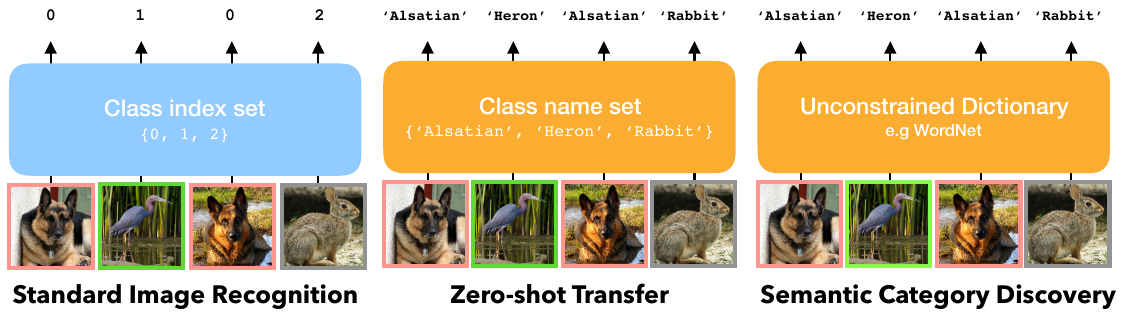}
    \caption{\textbf{An illustration of how our proposed tasks extend existing image recognition settings.} Left: A model is trained to predict class indices for a pre-defined set of categories (\eg, supervised recognition, unsupervised clustering, category discovery etc.) Middle: A vision-language model is given a pre-defined set of class names to recognize images in a `zero-shot' manner. Right: In our proposed setting, a model must predict an image's class name given only a large, unconstrained vocabulary. Note that the leftmost setting (blue model) uses only a visual representation, while the middle and right settings (orange models) use vision-language representations.}
    \label{fig:main}
\end{figure*}
\section{Introduction}
\label{sec:intro}
Image recognition has emerged as a fundamental task for demonstrating progress in computer vision and machine learning ~\cite{he2016deep, Simonyan15vgg, caron2020unsupervised, caron2021emerging, Radford2021Learning}. 
\jie{In this work, we bring the current image recognition pipeline closer to that of the human visual system.}
Specifically, given a set of images, humans are able to directly map them to semantic object names --- \eg, `that is a bird', `that is a fish', `that is a lion' if one were visiting a zoo. 
In a `zero-shot' fashion, a human can leverage relations between images to narrow down a large list of possible nouns to a precise set of semantic labels.

In contrast, consider the conventional image recognition task, which involves mapping a set of images to a static set of class indices ~\cite{ILSVRC15, Krizhevsky09cifar}, where these indices represent a predefined set of class names.
\jie{
One key limitation is the finite set of class labels provided to the model as priori information for recognition. 
In the open world, it is often difficult to have a complete label set in advance (\eg, automatic driving), and new categories keep emerging (\eg, products in the supermarket).}
The classification output is the category index, not the semantic vocabulary that humans can directly understand, so it needs to be mapped to the semantic vocabulary.
Meanwhile, recent vision-language models trained on internet scale data are endowed with `open' vocabularies~\cite{Radford2021Learning,Jia2021align}. 
These models have the ability to map images into a representation space where they can be directly compared with semantic labels. 
However, again, during evaluation, the models are always given a limited set of nouns from which to select the best match for an image.

This is the problem we tackle in this paper: given a collection of images and a large (essentially `open') vocabulary, assigning class names to each image.
\jie{We term this task \textit{Semantic Category Discovery}, or SCD.}
Our solution to this is to provide a vision-language model with a large, `open' vocabulary of possible concepts (e.g WordNet with $68k$ nouns), and use relations between images to find the most relevant set of concepts for the task.
Specifically, we propose a solution based on non-parametric clustering and iterative refinement.
We first group the images using off-the-shelf non-parametric clustering methods on top of self-supervised features, before using a vision-language model to infer an initial set of candidates' names from the open vocabulary. 
Using a voting strategy within clusters to narrow down the initial set of candidate names, followed by an iterative refinement process, we build a zero-shot classifier that can reliably predict the class names of images.
Additionally, 
as enhancing the visual feature with relevant textual features can effectively improve image clustering performance~\cite{text_guided_cluster,clipgcd}, we explore using vision-text features for SCD. 
Specifically, we leverage the CLIP vision and text encoders to retrieve the most relevant text descriptions for each image from an external database, \emph{i.e.}, Conceptual Captions 12M (CC12M) \cite{cc12m}. We can then use the vision-text features, instead of the vision-only features for the initial non-parametric clustering in our approach.

We evaluate our solution in a number of experimental settings.
We control for:
the initial choice of vocabulary, using the generic WordNet vocabulary as well as a task-specific vocabulary of $10k$ bird species; 
and also dataset granularity, including a large-scale ImageNet experiment ~\cite{ILSVRC15} as well as a fine-grained CUB evaluation ~\cite{WahCUB_200_2011}. 
We further conduct experiments when only images are available (the unsupervised setting) as well as when we have labels for a subset of the images (partially supervised ~\cite{vaze2022gcd}). 
For the latter, we further introduce a constrained semi-supervised clustering algorithm based on a Minimum Cost Flow (MCF) optimization problem ~\cite{bennett2000constrained}.
In all cases, we find that our method substantially improves performance over baselines (e.g 50\% relative improvement on ImageNet). 
In fact, surprisingly, we find that our method not only allows models to \textit{name} the images but in many cases, improves \textit{clustering performance} over \jie{the} prior state-of-the-art.

In summary, we make the following contributions: 
\textcolor{blue}{(i)} We propose a visual recognition task that is closely aligned to that tackled by humans, which involves assigning semantic label names to a collection of images, and which addresses limitations in current tasks; 
\textcolor{blue}{(ii)} We propose novel solutions to this task using non-parametric clustering and iterative refinement with pre-trained vision-language models; 
\textcolor{blue}{(iii)} In a range of evaluation settings, we find that our method achieves \jie{a} substantial improvement of existing baselines and prior state-of-the-art. The task we tackle, and its distinctions with prior problems, is illustrated in ~\cref{fig:main}.



\section{Related work}
\label{sec:related}

Our work concerns assigning semantic label names to a collection of unlabelled images.
Here, we review the two most closely related fields to this task: \textit{clustering/category discovery}, which aims to group sets of semantically related images together, and \textit{semantic representation learning}, which concerns joint learning of visual and semantic (text) representations.

\begin{figure*}
    \centering
    \includegraphics[width=0.8\linewidth]{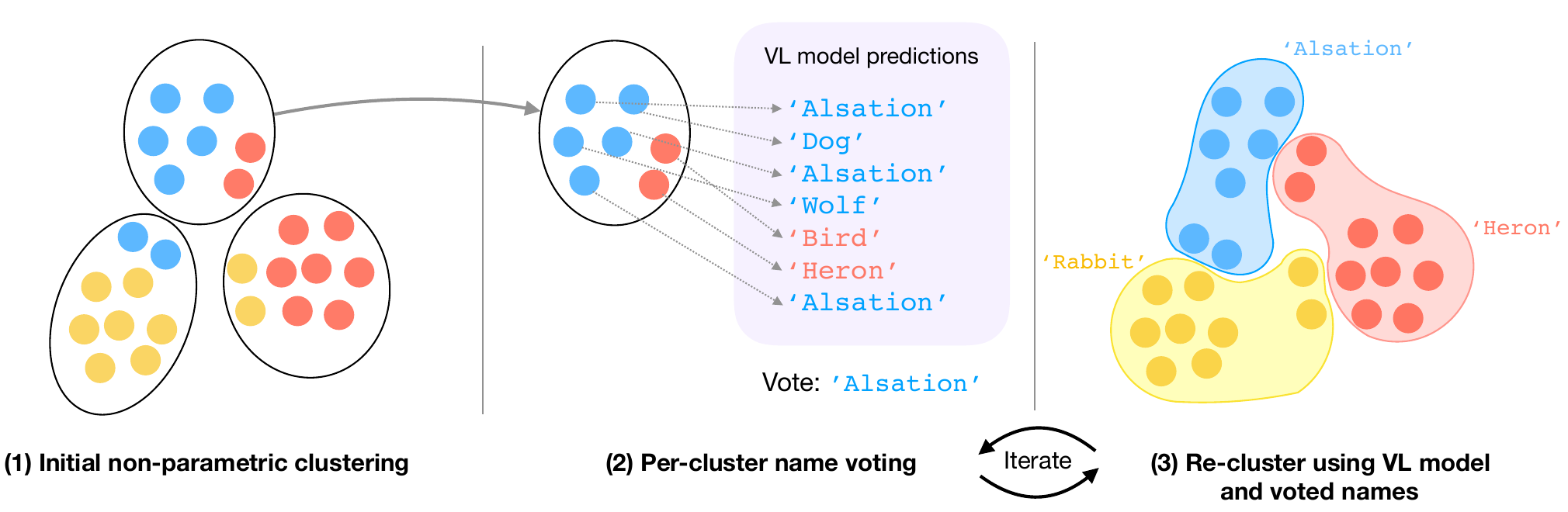}
    \caption{An illustration of our proposed method. Left: We first perform non-parametric clustering on deep features to get initial cluster assignments (see \cref{sec:method:unsupervised:init}). Middle: For each cluster, we use a VL model to assign a class name for each image from the \textit{entire open vocabulary}. We select one class name for each cluster based on the most common occurrence (see \cref{sec:method:name_voting}). Right: Based on the voted class names, we label each image as one of these, using these assignments to form new clusters. 
    We then iterate between name voting and re-clustering. 
    Note: Here, we have not illustrated refinement with top-$k$ voting (see \cref{sec:method:unsupervised:refine}).}
    \label{fig:method}
\end{figure*}

\vspace{-0.2cm}
\paragraph{Clustering and category discovery}
\textit{Unsupervised clustering} is a canonical problem in machine learning, where the aim is to find natural groupings of data without any labels ~\cite{Aggarwal13cluster, vangansbeke2020scan, MackQueen67_Kmeans, LLoyd_kmeans, yang_icml_17}.
With the advent of deep learning, many works have tackled the problem of simultaneously learning an embedding space and clustering data ~\cite{ji2019invariant, caron2018deep, Ronen_DeepDPM}. 
For instance, Invariant Information Clustering, IIC ~\cite{ji2019invariant}, seeks to learn a representation which maximizes the mutual information between class assignments of two augmentations of the same image, while DeepCluster ~\cite{caron2018deep} iteratively performs $k$-means ~\cite{MackQueen67_Kmeans, LLoyd_kmeans} and gradient descent, where the cluster assignments are used as pseudo-labels for learning.
Recent work has shown that, on top of well-trained features, the classic $k$-means algorithm can perform comparably to more sophisticated methods ~\cite{Ronen_DeepDPM}.

Meanwhile, a rich vein of research considers `category discovery', where labelled data is leveraged to discover and group images from new categories in unlabelled data.
\textit{Novel Category Discovery} (NCD) considers the setting in which the categories in the labelled and unlabelled data are disjoint ~\cite{han2019learning, zhao21novel, han20automatically,han21autonovel, jia21joint}. 
DTC~\cite{han2019learning} approaches this problem by first learning an embedding from the labelled data and using the representation to cluster unlabelled data, while ~\cite{Fini_2021_ICCV} proposes a SWaV-like ~\cite{caron2020unsupervised} clustering approach to partition the unlabelled data. Very recently, NCD has been extended to a more generalized setting as \textit{`Generalized Category Discovery'} (GCD)~\cite{vaze2022gcd} and concurrently as \textit{`Open-World Semi-supervised Learning'} ~\cite{cao2022openworld}, wherein unlabelled data can include instances from both \jie{predefined} and new categories.

All methods and tasks described here involve only grouping similar images together or, equivalently, predicting a cluster index for each image.
In contrast, we extend these works by directly predicting an interpretable, semantic \textit{class name}, which we suggest results in a more useful and practical perception system.

\vspace{-0.2cm}
\paragraph{Semantic representation learning}
Most image recognition models are trained on fully annotated data with a predefined set of class indices~\cite{NIPS2012AlexNet, Simonyan15vgg, he2016deep}. 
Here, semantics are only recovered by mapping indices to names, which is limited by the set of mappings defined \jie{a priori}.
Meanwhile, multi-modal settings with \textit{vision-language models} involve the learning of an embedding space where images and semantic text are represented jointly ~\cite{Frome2013devise,Karpathy2015,Faghri2018Faghri,Desai21virtex,Chen21learning,Radford2021Learning,Jia2021align}. 
Notably, contrastive learning with image-text pairs on internet-scale data (\eg, CLIP~\cite{Radford2021Learning} and ALIGN~\cite{Jia2021align}) results in a robust representation which embeds an `open' vocabulary of semantic concepts.
LiT~\cite{Zhai2022lit} also introduced a more data efficient image-text embedding approach by freezing a pre-trained image encoder and tuning only the text embedding. 
These large-scale models have been applied to a range of downstream tasks such as image captioning~\cite{Lu2019ViLBERT}, VQA~\cite{Goyal2017Making}, visual commonsense reasoning~\cite{Zellers2019From}, etc. 
They further show encouraging \textit{zero-shot transfer} performance, where the models are repurposed for image recognition without any labelled data in the target domain.
In this case, at test-time, the models are given a finite set of candidate class names from which to select the best match for a given image. Besides, a discussion on multi-modal large language models~\cite{llava, gpt4} is included in the supplementary material.

\jie{In this work, we leverage the vision-language models to explore a new fashion for classification and recognition tasks. The proposed method can facilitate automatic semantic category discovery from images with unconstrained vocabulary instead of with the closed-set ground truth.}

\vspace{-0.2cm}
\paragraph{Object discovery} 
Object discovery considers the problem of localization and/or segmenting the unlabelled objects from the queried Internet data under given classes (\eg, the image searching results of `Car’ by Google). Methods with varying supervision have been proposed to tackle this task, including weakly supervised \cite{choe2020cvpr,Deselaers2010eccv}, semi-supervised \cite{Chen2013NEIL,Chen2014enrich}, self-supervised \cite{Henaff22}, and unsupervised ones  \cite{vo19unsup,NEURIPS2021_8bf1211f}. 

\jie{In this work, we focus on discovering semantic categories and predicting the semantic names of the discovered categories, rather than localizing objects.}






\section{Semantic category discovery}
\label{sec:method}

\newcommand{\xh}[1]{\textcolor{green}{[Xh: #1]}}

In this paper, we consider the problem of Semantic Category Discovery. 
\jie{The pipeline of this task is shown in \cref{fig:main}.}
Given a collection of images, the objective is to automatically assign a class name to each image, with only a large (`open') vocabulary as prior information. 
In general, we assume datasets with $K$ distinct semantic categories, and the aim is to find the optimal set of $K$ category names and assign each image to one of them.

In this section, we first describe our baseline for this task in \cref{sec:zero_shot_baseline}, before describing our solutions to SCD in two settings: a fully unsupervised setting where we have no labelled data (\cref{sec:method:unsupervised}); and a partially supervised setting, similar to Generalized Category Discovery ~\cite{vaze2022gcd}, where we have some labelled data (though not from all categories, \cref{sec:method:partsupervised}).

\subsection{Baseline: zero-shot transfer}
\label{sec:zero_shot_baseline}

Recent large-scale vision-language (VL) models, such as CLIP~\cite{Radford2021Learning} and ALIGN~\cite{Jia2021align}, have demonstrated the ability to match images with arbitrary textual (semantic) inputs by modelling in a joint embedding space. 
These models are pre-trained with weak supervision on large-scale image-text pairs. 
In current evaluation settings, models are fed a finite set of $K$ possible nouns and tasked with selecting the best match for a given image.
For instance, when evaluating on ImageNet, VL models are given the class names corresponding to the $K=1000$ categories in the standard ILSVRC dataset, and the models are expected to select one out of the 1000 predefined true names describing all classes of ImageNet.
\textit{We argue that this is unrealistic prior knowledge to obtain for many real-world tasks, and that humans are able to narrow down a large, open vocabulary to assign semantic class names.
}

Our baseline for SCD respects this intuition. 
Specifically, instead of giving the VL model $K$ possible names for an image, we give it all nouns from a large dictionary, of size $N \gg K$, which is a proxy for all possible nouns.
This baseline is termed \textit{zero-shot transfer} in this work.
Unsurprisingly, if we relax the semantic space from the true set of categories to a universal space, we find the model's performance degrades substantially.
The goal in this work can be seen as finding a way to narrow the large dictionary of size $N$ to the ground truth set of $K$ class names \textit{automatically}.

\subsection{Unsupervised setting}
\label{sec:method:unsupervised}
We first consider the situation where we only have unlabelled images.
Different from conventional clustering, we wish to recognize the images by their semantic names rather than simply a cluster index.
Specifically, we consider a collection of unlabelled images $\mathcal{D_U} = \{(\mathbf{x}_i, y_i)\}_{i=1}^{M} \in \mathcal{X} \times \mathcal{Y_U}$, where $y_i \in \mathcal{Y_U} = \{1, \dots, K\}$ are the ground truth class indices associated with a unique noun.

Our solution here leverages a pre-trained VL model (\eg, CLIP~\cite{Radford2021Learning}) and non-parametric clustering. In a nutshell, we propose to first cluster images in an embedding space, before using the VL model to bridge the gap between cluster assignment and object semantics. 
Next, given an initial guess of the set of semantic category names, we perform iterative refinement to improve the predictions.

\subsubsection{Initial clustering} 
\label{sec:method:unsupervised:init}
Image recognition with standard VL models is a form of \textit{parametric classification}. 
Specifically, the embeddings of the candidate semantic names through a text encoder form a (parametric) linear classifier on top of visual features.
In contrast, we propose to also leverage a non-parametric recognition signal through the form of clustering. 
~\cite{Ronen_DeepDPM} shows that clustering of features is a good replacement for parametric classification in the case when we have limited labelled data; in our case, the clustering allows us to establish relationships \textit{between data points} which the VL model would otherwise be blind to.
In detail, given a feature extractor $f_{\theta}$,  we extract a feature vector $\mathbf{z}_i$ as $\mathbf{z}_i=f_{\theta}(\mathbf{x}_i)$ for each image $\mathbf{x}_i$ in $\mathcal{D_U}$. 
We then run the standard $k$-means algorithm to partition the data into $K$ clusters. 

\subsubsection{Name voting}
\label{sec:method:name_voting}
Next, we use the cluster assignments to narrow down a large, open dictionary to an initial estimate of the set of $K$ class names by voting on class names within a cluster. 
In particular, provided with the cluster assignment $y'_i$ for each $\mathbf{x}_i$, we collect instances in a cluster with index $c$ as $\mathcal{D}_c = \{\mathbf{x}_i \ | \ y'_i = c, \mathbf{x}_i \in \mathcal{D_U}\} $. 
Let $f_{\omega}^v$ and $f_{\omega}^t$ be the pre-trained vision and text encoders of a large VL model. 
We extract visual features $\mathcal{Z}_\mathcal{V}^c = \{\mathbf{z}_i^v = f_{\omega}^v(\mathbf{x}_i) \ | \ \mathbf{x}_i \in \mathcal{D}_c\}$, and text embeddings for all nouns in the large dictionary $\mathcal{N}$ as $\mathcal{S_T} = \{\mathbf{s}_i^t = f_{\omega}^t(\mathbf{n}_i) \ | \ \mathbf{n}_i \in \mathcal{N}\}$. For each $\mathbf{z}_i^v$ in cluster $c$, a predicted semantic name $\mathbf{s}_{\alpha(i)}$ can be obtained by querying the nearest neighbour (NN) in $\mathcal{S_T}$:

\begin{equation}
    \alpha(i) = \mathop{\arg\max}_{j} \ \{\mathbf{z}_i^v \cdot\mathbf{s}_j^t \ |\ \mathbf{s}_j^t \in \mathcal{S_T}\}.
    \label{eq:nn}
\end{equation}

After obtaining predicted the semantic name embedding vector $\mathbf{s}_{\alpha(i)}$ for each image $i$ in cluster $c$, all the unique text embeddings in $\{\mathbf{s}_{\alpha(i)}\}$ forms a subset of  $\mathcal{S_T}$, denoted as $\mathcal{S}_\mathcal{T}^c$.
We simply choose the most common semantic name in the cluster to give an initial set of candidate names $\mathbf{S}_\mathcal{T}^\mathcal{D_U}$ as: 
\begin{equation}
\mathbf{S}_\mathcal{T}^\mathcal{D_U} = \{ \mathbf{s}_c =  \mathop{\arg\max}_{\mathbf{s}_j}\mathcal{P}_c(\mathbf{s}_j) \ | \ \mathbf{s}_j \in \mathcal{S}_\mathcal{T}^c, c = 1, \cdots, K\},
\label{eq:stdu}
\end{equation}
where $\mathcal{P}_c(\mathbf{a})$ counts the occurrences of the semantic vector $\mathbf{a}$ of a noun in cluster $c$.

\subsubsection{Semantic refinement}
\label{sec:method:unsupervised:refine}

Having obtained an initial estimate of the set of $K$ class names, we perform two \textit{refinement} steps.
Firstly, we note that by enforcing one semantic name to represent each cluster in the initial clustering step, any duplicated class names would necessarily give us a smaller set of candidate names than those actually present in $\mathcal{D_U}$.
To remedy this problem, instead of measuring the frequency of only the nearest semantic name in~\cref{eq:nn} for each instance in a cluster $c$, we instead track the frequency of the $m$-NN. 
In this way, for each cluster $c$, there are $m$ candidate semantic vectors. 
This allows us to form a cost matrix of assigning a class name to a cluster, where the cost is equal to the proportion of instances in that cluster assigned that class name.
Next, we solve the linear assignment problem between class names and clusters, using the Hungarian algorithm to assign one semantic vector to each cluster without duplication. 

At this point, by stacking all the semantic vectors in $\mathbf{S}_\mathcal{T}^\mathcal{D_U}$, we construct a $d\times K$ matrix forming a parametric, zero-shot linear classifier $\phi :\mathbb{R}^d \rightarrow\mathbb{R}^K$. 
By applying this classifier on the visual features as $j' = \mathop{\arg\max}_{j} \phi(\mathbf{z}_i^v)[j]$, we compute semantic assignments and can also construct a new clustering of the data, which is not necessarily the same as the initial clustering assignment achieved with the non-parametric algorithm.
This new clustering of the data allows us to \textit{iteratively update} the candidate semantic vectors and the cluster assignments until convergence while ensuring $K$ unique semantic class names are assigned.

Intuitively, our final iterative process \jie{is analogous to the classic $k$-means algorithm.} 
Clusters are iteratively created and updated based on energy. However, the energy in our case is based on similarities of the visual features to a set of semantic text embeddings rather than similarities between visual features (as in standard $k$-means).

\subsection{Partially supervised setting}
\label{sec:method:partsupervised}

We also consider a partially supervised setting where, in addition to the collection of unlabelled images $\mathcal{D_U} = \{(\mathbf{x}_i, y_i)\}_{i=1}^{M} \in \mathcal{X} \times \mathcal{Y_U}$, we also have access to some labelled images  $\mathcal{D_L} = \{(\mathbf{x}_i, y_i)\}_{i=1}^{N} \in \mathcal{X} \times \mathcal{Y_L}$, where $\mathcal{Y_L} \subset \mathcal{Y_U}$.
This setting has recently been formalized as `Generalized Category Discovery' \cite{vaze2022gcd}, which we now extend to require the prediction of semantic names rather than simply class indices.

Here, one could simply apply the method from \cref{sec:method:unsupervised}, and ignore the labelled data. 
However, we can also leverage the labelled data to help us better recognize the unlabelled images. 
Similarly to~\cref{sec:method:unsupervised}, we first obtain cluster indices through a non-parametric clustering algorithm, followed by narrowing down the candidate semantic names from the universal semantic space to a set of elements with the same or more semantic names than the actual class number. 
However, differently to~\cref{sec:method:unsupervised}, we constrain the clustering stage using the labelled set.

\vspace{-0.2cm}
\subsubsection{Constraining k-means}
\label{sec:method:partsupervised:init}
With $\mathcal{D} = \mathcal{D_L}\cup  \mathcal{D_U}$, we adopt a semi-supervised $k$-means algorithm (SS-$k$-means)~\cite{han2019learning,vaze2022gcd} to obtain the cluster indices for each image. 
SS-$k$-means extends $k$-means to cluster partially labelled data $\mathcal{D}$ by enforcing labelled instances (in $\mathcal{D_L}$) of the same class to fall into the same cluster. 
Though it performs well in the GCD setting, we find that SS-$k$-means often produces clusters that contain very few instances.
This is the result of the supervision, as we find that the forced assignment on the labelled set results in a few `attractor' clusters which are bloated in size. 
Here, we improve SS-$k$-means by introducing a loose cluster size constraint. 
Specifically, we constrain the minimum size of the clusters during the SS-$k$-means clustering. 
Drawing inspiration from~\cite{bennett2000constrained}, we formulate the cluster assignment step of SS-$k$-means as a Minimum Cost Flow (MCF) linear optimization problem. 
We call the resulting algorithm constrained SS-$k$-means (CSS-$k$-means). 
After running CSS-$k$-means on $\mathcal{D}$, we obtain initial cluster assignments $y_i'$ for each $\mathbf{x}_i$ in the unlabelled data $\mathcal{D_U}$. The refinement process remains largely the same as in \cref{sec:method:unsupervised:refine}.
However, the search space for the vision language model is reduced, as we know that categories in $\mathcal{D_L}$ must be present.

\subsection{Enhancing the clustering feature with text}
\label{sec:method:textembedding}
\begin{figure}
    \centering
    \includegraphics[width=0.85\linewidth]{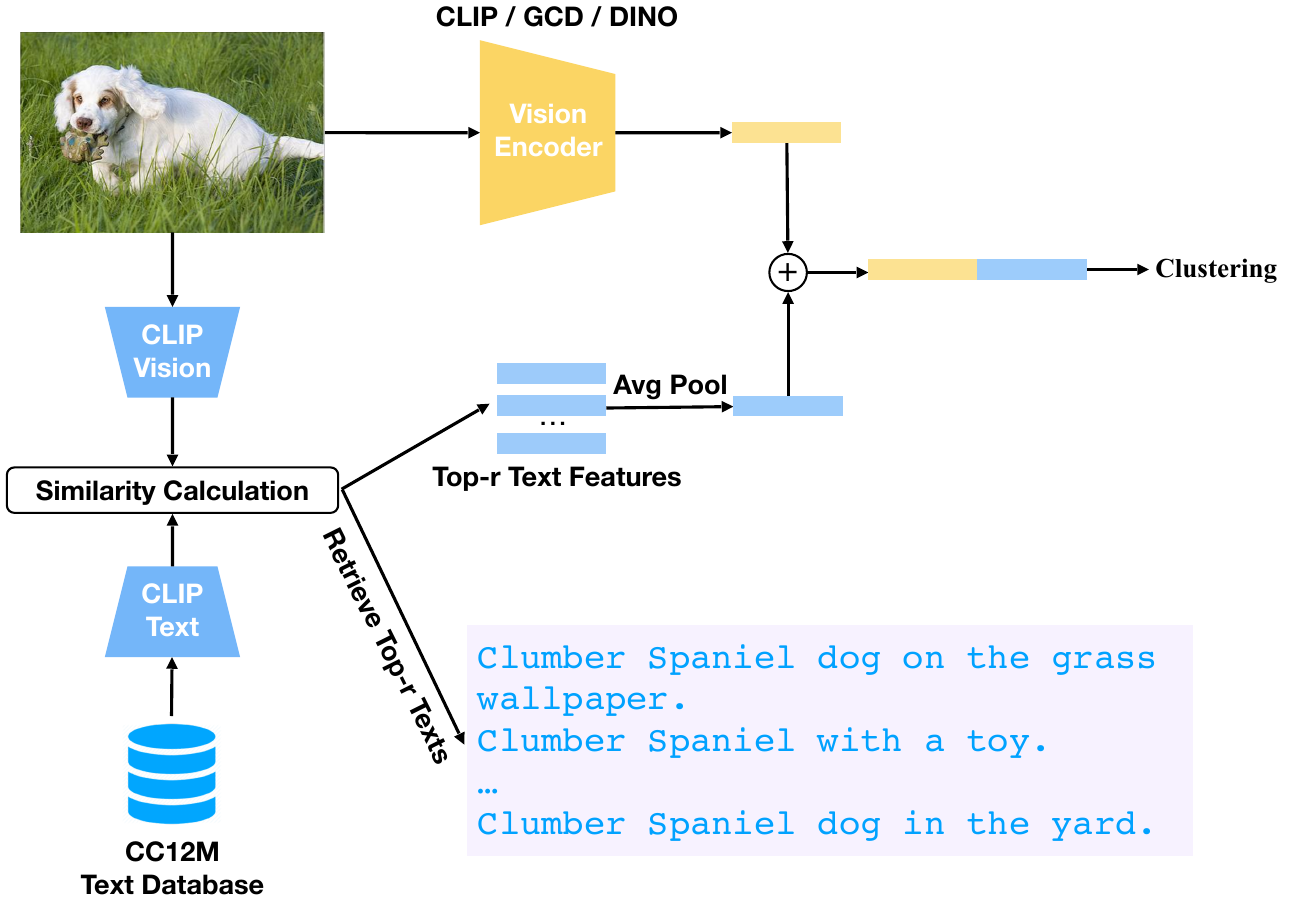}
    \caption{\textbf{Enhancing the clustering feature with text.}
    We retrieve the top-$r$ most relevant texts from the CC12M dataset \cite{cc12m} for each image using the CLIP vision and text encoders. 
    The textual features of the retrieved texts are combined with the visual features (obtained by a pre-trained vision encoder such as CLIP, DINO, and GCD) for the non-parametric clustering.}
    \label{fig:text_retrieval}
\end{figure}
For the initial non-parametric clustering, instead of using the vision-only features, we can also enhance the visual feature by leveraging the textual features of the relevant text from an external text database. 
Particularly, as illustrated in \cref{fig:text_retrieval}, for each image, we retrieve the top-$r$ most relevant texts from CC12M \cite{cc12m}, using the CLIP vision and text encoders. Specifically, we measure the cosine similarity between the image feature and all text features in the database and select the top-$r$ text features with the highest similarity to the image. The retrieved text features are then fused using average pooling. 
The fused textual feature is then combined (\eg, through concatenation along the channel dimension) with the visual feature obtained using a vision encoder (which can be CLIP, DINO or GCD pre-trained encoder, depending on the experimental setting). 
The resulting vision-text features are then used for the initial non-parametric clustering.
\section{Experiments}
\label{sec:exp}

\subsection{Vocabularies and datasets}

Recall that our aim is to take a collection of images and, using a VL model, assign a semantic class name to every image, given only a large vocabulary as prior information.
As such, a key component in our pipeline is the choice of vocabulary.
In this work, we consider nouns from the generic WordNet taxonomy, which contains $68k$ concepts and can be considered to contain most objects a human might typically encounter.
We further test a domain-specific vocabulary for naming bird species, for which we scrape $11k$ bird names from Wikipedia.

For evaluation data, we first demonstrate results on ImageNet \cite{ILSVRC15}, where each class name is a node in the WordNet hierarchy.
For comparison with GCD models ~\cite{vaze2022gcd}, we demonstrate results on ImageNet-100, a 100 class subset of the standard ILSVRC benchmark ~\cite{ILSVRC15}.
We also show results on the full dataset in ~\cref{tab:imagenet_1k_main}.
Furthermore, we show results on \textit{Stanford Dogs} ~\cite{KhoslaYaoJayadevaprakashFeiFei_FGVC2011}, a fine-grained dataset of 120 dog breeds, whose classes also come from the WordNet hierarchy. 
Finally, for an evaluation with the bird species vocabulary, we experiment with CUB ~\cite{WahCUB_200_2011}.
In the partially supervised setting, we use the data splits from GCD and refer to the original paper for details.


\subsection{Evaluation metrics}
\label{sec:metrics}

The nature of the Semantic Category Discovery task can be captured by the metrics we choose for evaluation.
The primary aim of this work is assigning \textit{semantic names} to a given image, rather than a typical class index as in standard recognition.
To this end, we first measure `Semantic Accuracy' or `sACC', which is analogous to standard classification accuracy. 
Specifically, given the ground truth semantic name, $s_i$, and the predicted name, $\hat{s}_i$:

\begin{equation}
    sACC = 
    \frac{1}{M}
    \sum_{i=1}^{M}
    \mathds{1}\{s_i = \hat{s}_i\},
\end{equation}

\begin{table*}[t]
\footnotesize
\caption{\textbf{Results in the unsupervised setting.} We use DINO features for the initial clustering step and report metrics for semantic accuracy (involving class naming, left) and clustering (right). `TE' denotes using the textual enhancement technique.}
\centering
\resizebox{.53\textwidth}{!}{
\begin{tabular}{lcccccc}
\toprule
& \multicolumn{2}{c}{ImageNet-100} & \multicolumn{2}{c}{Stanford Dogs} 
& CUB
\\
\cmidrule(rl){2-3}
\cmidrule(rl){4-5}
\cmidrule(rl){6-6}
& sACC &   Soft-sACC   
& sACC &    Soft-sACC  
& sACC
\\
\midrule
\rowcolor{Gray}
Zero-shot transfer (UB) 
& 85.0 & 92.0
& 60.4 & 83.2
& 54.1
\\
\midrule
Zero-shot transfer (Baseline) 
& 22.7 & 57.7
& 51.7 & 77.4
& 20.2
\\
\midrule
Ours (Semantic Naming) 
& 41.2 & 71.3
& 53.8 & 79.1
& 24.5
\\
\midrule
Ours (Semantic Naming) \textit{w/}{TE}
& \textbf{43.0} & \textbf{72.5}
& \textbf{54.1} & \textbf{80.0}
& \textbf{33.5}
\\
\bottomrule
\end{tabular}
}
\resizebox{.44\textwidth}{!}{
\begin{tabular}{lccc}
\toprule
& ImageNet-100 & Stanford Dogs 
& CUB
\\
\cmidrule(rl){2-2}
\cmidrule(rl){3-3}
\cmidrule(rl){4-4}
& ACC & ACC & ACC
\\
\midrule
\rowcolor{Gray}
Zero-shot transfer (UB) 
& 85.1 & 60.8 & 55.8
\\
\midrule
$k$-means (Baseline) 
& 73.2 & 47.2 & 34.4
\\
\midrule
Ours (Semantic Naming) 
& 78.2 & 57.9 & \textbf{46.5}
\\
\midrule
Ours (Semantic Naming) \textit{w/}{TE}
& \textbf{81.3} & \textbf{58.7}
& 42.6
\\
\bottomrule
\end{tabular}
}
\label{tab:unsup_main}
\end{table*}
\begin{table*}[!htb]
\footnotesize
\caption{\textbf{Results in the partially supervised setting.} We use GCD features for the initial clustering step and report metrics for semantic accuracy (involving class naming, left) and clustering (right). `TE' denotes using the textual enhancement technique.}
\centering
\resizebox{.53\textwidth}{!}{
\begin{tabular}{lcccccc}
\toprule
& \multicolumn{2}{c}{ImageNet-100} & \multicolumn{2}{c}{Stanford Dogs} 
& CUB
\\
\cmidrule(rl){2-3}
\cmidrule(rl){4-5}
\cmidrule(rl){6-6}
& sACC &   Soft-sACC   
& sACC &    Soft-sACC  
& sACC
\\
\midrule
\rowcolor{Gray}
Zero-shot transfer (UB) 
& 85.0 & 92.0
& 60.4 & 83.2
& 54.1
\\
\midrule
Zero-shot transfer (Baseline) 
& 22.7 & 57.7
& 51.7 & 77.4
& 20.2
\\
\midrule
Ours (Semantic Naming) 
& 54.8 & \textbf{77.5}
& 53.7 & 79.6
& 28.2
\\
\midrule
Ours (Semantic Naming) \textit{w/}{TE}
& \textbf{55.7} & 76.5
& \textbf{55.5} & \textbf{80.6}
& \textbf{35.3}
\\
\bottomrule
\end{tabular}
}
\resizebox{.41\textwidth}{!}{
\begin{tabular}{lccc}
\toprule
& ImageNet-100 & Stanford Dogs 
& CUB
\\
\cmidrule(rl){2-2}
\cmidrule(rl){3-3}
\cmidrule(rl){4-4}
& ACC & ACC & ACC
\\
\midrule
\rowcolor{Gray}
Zero-shot transfer (UB) 
& 85.1 & 60.8 & 55.8
\\
\midrule
GCD \cite{vaze2022gcd} (Baseline) 
& 74.1 & 60.8 & \textbf{54.0}
\\
\midrule
Ours (CSS-$k$-means) 
& 78.7 & \textbf{62.1} & 52.9
\\
Ours (Semantic Naming) 
& \textbf{83.0} & 56.6 & 46.8
\\
\midrule
Ours (Semantic Naming) \textit{w/}{TE}
& 80.6 & 58.8
& 42.5
\\
\bottomrule
\end{tabular}
}
\label{tab:partially_sup_main}
\end{table*}

Secondly, we introduce a soft evaluation metric which accounts for the continuous nature of semantic similarity.
For instance, if a `mushroom' is predicted as `fungus', the prediction should not be considered as completely wrong.
It would be preferable to account for semantic similarity between the ground-truth and predicted name during evaluation and, as such, we introduce the `soft Semantic Accuracy (Soft-sACC) metric. 
We adopt the Leacock Chodorow Similarity (LCS)~\cite{wordnet}, which measures lexical semantic similarity by finding the shortest path in the WordNet graph between two concepts, and scales that value by the maximum path length. 
The LCS between two concepts $i$ and $j$ is defined as:
\begin{equation}
   s_{i,j} = -log(\frac{p(i, j)}{2d}),
\end{equation}

where $p(i, j)$ denotes the shortest path length between $i$ and $j$ and $d$ denotes the taxonomy depth. 
We further re-scale the similarity score into the range $[0, 1]$ by dividing with the maximum possible score, with `0' indicating no semantic similarity and `1' indicating a perfect prediction. We evaluate the Soft-sACC on datasets for which the concepts are determined in the WordNet hierarchy: ImageNet-100, Standord Dogs, and ImageNet-1K.

Finally, as in the existing recognition literature, we are also interesting in \textit{clustering performance}, or how often images from the same category are grouped together.
A model could predict the wrong name for every image but still achieve high `Clustering Accuracy' (ACC), as long as images from the same category were predicted the same (incorrect) name. 
Clustering Accuracy is defined as:
\begin{equation}
    ACC = \max_{p \in \mathcal{P(Y_U)}}
    \frac{1}{M}
    \sum_{i=1}^{M}
    \mathds{1}\{y_i = p(\hat{y}_i)\},
\end{equation}
where $y_i$ represent the ground truth label and $\hat{y}_i$ the cluster assignment and $\mathcal{P(Y_U)}$ denotes all possible permutations of the class labels in the unlabelled data. 
Note that `sACC' is generally a more difficult metric than `ACC', with random guessing performance being roughly $1/N$ (\eg, $1/68k$ for WordNet), while random guessing for `ACC' is roughly $1/K$ (\eg, $1/100$ for ImageNet-100).

\subsection{Implementation details}
\par\noindent\textbf{Models.} For the visual feature extractor $f_\theta$ used for initial cluster assignment (\cref{sec:method:unsupervised:init} 
) we could, in principle, use any feature extractor. 
For the unsupervised setting, we use a self-supervised vision transformer (DINO ViT-B-16 weights ~\cite{Dosovitskiy2021,caron2021emerging}), while in the partially supervised setting, we fine-tune the feature extractor with supervised and self-supervised contrastive losses as in ~\cite{vaze2022gcd}.
For the vision-language component ($f_{\omega}^v$, $f_{\omega}^t$), we adopt the off-the-shelf pre-trained CLIP model with a ViT-B-16 backbone.
With the pre-trained CLIP model, we can retrieve top-10 text descriptions from the CC12M dataset to enhance the visual features for the initial non-parametric clustering.

\par\noindent\textbf{Compared methods.}
We can find no single existing method which can perform the Semantic Category Discover task and, as such, we compare against a number of methods to benchmark different aspects of our model. 
Firstly, we use the \textit{Zero-shot transfer (Baseline)} described in \cref{sec:zero_shot_baseline}.
Next, in the unsupervised setting, we employ \textit{$k$-means} on top of DINO features as in ~\cite{vaze2022gcd} to compare clustering accuracy. 
It has been shown in ~\cite{Ronen_DeepDPM} that $k$-means is competitive with complex state-of-the-art methods when the underlying features are good. 
In the partially supervised setting, we compare against \textit{GCD} ~\cite{vaze2022gcd}. 
We also provide an upper-bound for performance as \textit{Zero-shot transfer (UB)}.
Here, we evaluate performance if a zero-shot classifier is evaluated on the unlabelled data using only the ground class names.
Finally, we report our results as \textit{Ours (Semantic Naming)}.

\par\noindent\textbf{Results with an estimated number of categories} In the partially supervised case, we also report ACC after the initial non-parametric clustering step as \textit{Ours (CSS-KM)} to demonstrate the efficacy of our proposed constrained semi-supervised $k$-means algorithm.
Note that, in all cases, we assume knowledge of the ground-truth number of categories, $k$.
We highlight that the estimation of the true number of clusters in a dataset is its own challenging research problem ~\cite{Ronen_DeepDPM}.
As such, we treat the problem of estimating the number of categories as orthogonal to our semantic naming question. To deal with the cases with unknown $k$, we employ an off-the-shelf $k$-estimation technique from ~\cite{vaze2022gcd} to estimate $k$, and substitute this number into our method for semantic category discovery, as will be seen in~\cref{sec:estimated_k}. 

\begin{table}[!htb]
\footnotesize
\caption{\textbf{Results on the ImageNet-1k test set.} We evaluate on the standard ILSVRC ~\cite{ILSVRC15} benchmark in both unsupervised and partially supervised settings. We use DINO features to provide initial cluster assignments in this case. `TE' denotes using the textual enhancement technique.}
\centering
\resizebox{1.\linewidth}{!}{ 
\begin{tabular}{lccccccc}
\toprule
& \multicolumn{3}{c}{Unsupervised} & \multicolumn{3}{c}{Partially Supervised} 
\\
& sACC &   Soft-sACC  & ACC
& sACC &   Soft-sACC  & ACC
\\
\midrule
\rowcolor{Gray}
Zero-shot transfer (UB) 
& 63.4 & 81.3 & 64.1 
& 63.4 & 81.3 & 64.1
\\
\midrule
$k$-means 
& - & - & 50.2 
& - & - & 50.2  
\\
Zero-shot transfer (Baseline) 
& 24.3 & 57.5 & - 
& 24.3 & 57.5 & - 
\\
\midrule
Ours (Semantic Naming) 
& 31.1 & 63.5 & 55.5
& 37.7 & \textbf{66.7} & \textbf{54.9} 
\\
\midrule
Ours (Semantic Naming) \textit{w/}{TE}
& \textbf{32.7} & \textbf{64.2}
& \textbf{57.6} & \textbf{38.7}
& {66.6} & {54.0}
\\
\bottomrule
\end{tabular}
}
\label{tab:imagenet_1k_main}
\end{table}

Finally, no existing method can be evaluated in all settings. 
Specifically, clustering based methods cannot be evaluated for `sACC'.
Also, we found it computationally infeasible to compute the `ACC' for the zero-shot transfer baseline.
The baseline predicts too many unique class names (usually over 10k) to reasonably compute the Hungarian assignment with the ground truth classes. 
In \cref{tab:imagenet_1k_main}, when a method cannot be evaluated in a given setting, we fill the table entry with `-'.

\subsection{Main results}
We evaluate on ImageNet-100, Stanford Dogs and CUB in the unsupervised (\cref{tab:unsup_main}) and partially supervised (\cref{tab:partially_sup_main}) settings, respectively, providing results for both semantic namings (left-hand tables) and clustering performance (right-hand). 
We first note that the sACC of the baseline is surprisingly strong, given the random guessing performance of less \jie{than} 0.1\% in all cases. This speaks to the strength of the underlying large-scale VL model.
However, our method provides improvements on this across the board.
Specifically, our method roughly doubles sACC on ImageNet-100 in both cases and improves performance on CUB by roughly 50\% in the partially supervised setting for CUB. 
We note that gains are relatively modest for Stanford Dogs and in CUB's unsupervised evaluation. 
Next, we highlight a surprising finding that our method can usually also aid clustering accuracy. 
That is, in addition to our method giving the model the ability to assign a \textit{semantic name} to an image, the resulting classifications can also be used to cluster images more reliably than prior methods.

\begin{table*}[t]
\footnotesize
\caption{\textbf{Results in the partially supervised setting with estimated class numbers.} We use GCD features for the initial clustering step and report metrics for semantic accuracy (involving class naming, left) and clustering (right).}
\centering
\resizebox{.5\textwidth}{!}{
\begin{tabular}{lcccccc}
\toprule
& \multicolumn{2}{c}{ImageNet-100} & \multicolumn{2}{c}{Stanford Dogs} 
& CUB
\\
\cmidrule(rl){2-3}
\cmidrule(rl){4-5}
\cmidrule(rl){6-6}
& sACC &   Soft-sACC   
& sACC &    Soft-sACC  
& sACC
\\
\midrule
\rowcolor{Gray}
Zero-shot transfer (UB) 
& 85.0 & 92.0
& 60.4 & 83.2
& 54.1
\\
\midrule
Zero-shot transfer (Baseline) 
& 22.7 & 57.7
& \textbf{51.7} & 77.4
& 20.2
\\
\midrule
Ours (Semantic Naming) 
& \textbf{54.5} & \textbf{75.4}
& 49.7 & \textbf{78.1}
& \textbf{29.8}
\\
\bottomrule
\end{tabular}
}
\resizebox{.42\textwidth}{!}{
\begin{tabular}{lccc}
\toprule
& ImageNet-100 & Stanford Dogs 
& CUB
\\
\cmidrule(rl){2-2}
\cmidrule(rl){3-3}
\cmidrule(rl){4-4}
& ACC & ACC & ACC
\\
\midrule
\rowcolor{Gray}
Zero-shot transfer (UB) 
& 85.1 & 60.8 & 55.8
\\
\midrule
GCD \cite{vaze2022gcd} (Baseline) 
& 74.1 & 60.8 & \textbf{54.0}
\\
\midrule
Ours (CSS-$k$-means) 
& 75.5 & \textbf{60.9} & 52.2
\\
Ours (Semantic Naming) 
& \textbf{79.1} & 54.8 & 49.7
\\
\bottomrule
\end{tabular}
}
\label{tab:unsup_softsacc_unknow_class_number}
\end{table*}

Finally, we demonstrate results in both the unsupervised and partially supervised settings on a large-scale ImageNet evaluation in \cref{tab:imagenet_1k_main}.
Particularly, in the partially supervised case, we can substantially improve upon the baseline here, with a 13\% boost in sACC.

Furthermore, we assess the efficacy of incorporating vision-text features for clustering. As evident from \cref{tab:unsup_main}, \cref{tab:partially_sup_main}, and \cref{tab:imagenet_1k_main}, the integration of textual features for clustering yields performance improvements in most cases, thereby demonstrating the complementary information provided by the retrieved texts to visual cues. Moreover, we observe a general trend wherein the use of texts tends to achieve more significant performance gains in the unsupervised setting compared to the partially supervised setting. 
The reason is likely that, in the unsupervised setting, textual features can help bridge the gap when vision-only features are not sufficiently discriminative, while for partial supervision, labeled data offers a relatively strong constraint on the clustering process, restricting the impact of textual features.


\subsection{Unknown number of categories}
\label{sec:estimated_k}
When the category number is unknown, we use the method proposed in~\cite{vaze2022gcd} to estimate the number of categories in the unlabeled data for the partially supervised setting, and we get 109 classes for ImageNet-100, 114 classes for Stanford Dogs and 231 classes for CUB,  respectively.
The results are shown in~\cref{tab:unsup_softsacc_unknow_class_number}.
We find a reduction in performance on a number of datasets, including on Stanford Dogs.
However, on some datasets, we surprisingly find improved performance, which we attribute to a quirk of the specific interaction of the initial clustering and semantic naming process.
Note that, overall, our method still substantially outperforms baselines and prior art.


\subsection{Analysis}

Here we investigate the various design choices in our solution to SCD. In \cref{tab:unsup_compare}, we investigate both the choice of features for the initial clustering step, as well as the effects of different components of our semantic naming method, under the unsupervised setting. 
Particularly, we gradually add our initial voting step, the iterative refinement, and the linear assignment, as introduced in~\cref{sec:method:unsupervised:init} and \cref{sec:method:unsupervised:refine}, on top of the $k$-means non-parametric clustering. We experiment on the generic classification dataset ImageNet-100 and also the fine-grained Stanford Dogs dataset. 
We experiment with both CLIP~\cite{Radford2021Learning} and DINO~\cite{caron2021emerging} features for the initial $k$-means clustering.

\begin{table}[t]
\footnotesize
\caption{\textbf{Effectiveness of different components of our semantic naming method.} Experiments under the unsupervised setting with CLIP and DINO features.}
\centering
\resizebox{1.\linewidth}{!}{
\begin{tabular}{lcccccccc}
\toprule
& \multicolumn{4}{c}{CLIP} & \multicolumn{4}{c}{DINO} 
\\
\cmidrule(rl){2-5}
\cmidrule(rl){6-9}
Datasets       & \multicolumn{2}{c}{ImageNet-100}      & \multicolumn{2}{c}{Stanford Dogs}       & \multicolumn{2}{c}{ImageNet-100}      & \multicolumn{2}{c}{Stanford Dogs}       
\\
\cmidrule(rl){2-3}
\cmidrule(rl){4-5}
\cmidrule(rl){6-7}
\cmidrule(rl){8-9}
& sACC &   ACC & sACC &   ACC & sACC &   ACC & sACC &   ACC \\
\midrule
$k$-means & - & 62.3   & - &  26.9 & -   &  73.2 & -    & 47.2 \\
+ Initial Voting Step & 35.6 & 70.8    &  41.8 & 47.0     &  40.7 & 77.5     & 45.3 & 50.1\\
+ Iteration & 36.4 & 70.9    &  44.2 & 49.1 & 40.9 & 78.0     & 48.3 & 52.1\\
+ Linear assignment &  \textbf{37.1} & \textbf{71.3}  & \textbf{50.3} & \textbf{55.2}   & \textbf{41.2} & \textbf{78.2}   & \textbf{53.8} & \textbf{57.9} \\
\bottomrule
\end{tabular}
}
\label{tab:unsup_compare}
\end{table}

As can be seen, each component of the semantic naming process improves sACC and ACC. 
The trend holds for different datasets and features used for the initial clustering.
Furthermore, DINO features provide a better starting point than CLIP in this setting.
This is likely because DINO features form strong nearest-neighbour classifiers ~\cite{caron2021emerging}.
Our entire method gives an sACC of $41.2\%$ on ImageNet-100 and $53.8\%$ on Stanford Dogs using DINO based initial clustering.  
We also note the boost our methods can give in clustering accuracy on ImageNet-100 over $k$-means clustering with DINO features. 
This is surprising as, despite not being designed for this purpose, our semantic naming method is able to group similar images together more effectively than the strong DINO baseline.
We also investigate the effects of using different initial clustering algorithms. Compared with $k$-means and SS-$k$-means, the proposed CSS-$k$-means gives the best semantic naming performance. 

Please refer to the supplementary material for more experimental details and qualitative analysis.

\section{Conclusions}
\label{conclusion}
In this work, we have proposed and tackled the task of \textit{Semantic Category Discovery} (SCD), where a model must predict the \textit{semantic category name} of an image given only a large, open dictionary as prior information. 
This extends standard image recognition settings which only require mapping images to a pre-defined set of class indices. 
It also extends zero-shot evaluations of large-scale vision-language models, which assume a finite set of candidate class names will be given at \jie{test time}. 
We propose a solution based on parametric clustering followed by iterative semantic refinement.
Though conceptually simple and computationally inexpensive (requiring no training),
we demonstrate that this method substantially outperforms existing baselines for SCD on coarse and fine-grained datasets, including a full ImageNet evaluation.
We suggest that our resulting models are more human-compatible and also align better with \jie{the} capabilities of human perceptual systems.

\newpage

{\small
\bibliographystyle{ieeenat_fullname}
\bibliography{egbib}

\begin{thebibliography}{54}
\providecommand{\natexlab}[1]{#1}
\providecommand{\url}[1]{\texttt{#1}}
\expandafter\ifx\csname urlstyle\endcsname\relax
  \providecommand{\doi}[1]{doi: #1}\else
  \providecommand{\doi}{doi: \begingroup \urlstyle{rm}\Url}\fi

\bibitem[Achiam et~al.(2023)Achiam, Adler, Agarwal, Ahmad, Akkaya, Aleman, Almeida, Altenschmidt, Altman, Anadkat, et~al.]{gpt4}
Josh Achiam, Steven Adler, Sandhini Agarwal, Lama Ahmad, Ilge Akkaya, Florencia~Leoni Aleman, Diogo Almeida, Janko Altenschmidt, Sam Altman, Shyamal Anadkat, et~al.
\newblock Gpt-4 technical report.
\newblock \emph{arXiv preprint arXiv:2303.08774}, 2023.

\bibitem[Aggarwal and Reddy(2013)]{Aggarwal13cluster}
Charu~C. Aggarwal and Chandan~K. Reddy.
\newblock \emph{Data Clustering: Algorithms and Applications}.
\newblock CRC Press, 2013.

\bibitem[Bradley et~al.(2000)Bradley, Bennett, and Demiriz]{bennett2000constrained}
Paul~S Bradley, Kristin~P Bennett, and Ayhan Demiriz.
\newblock Constrained k-means clustering, 2000.

\bibitem[Cao et~al.(2022)Cao, Brbic, and Leskovec]{cao2022openworld}
Kaidi Cao, Maria Brbic, and Jure Leskovec.
\newblock Open-world semi-supervised learning.
\newblock In \emph{ICLR}, 2022.

\bibitem[Caron et~al.(2018)Caron, Bojanowski, Joulin, and Douze]{caron2018deep}
Mathilde Caron, Piotr Bojanowski, Armand Joulin, and Matthijs Douze.
\newblock Deep clustering for unsupervised learning of visual features.
\newblock In \emph{ECCV}, 2018.

\bibitem[Caron et~al.(2020)Caron, Misra, Mairal, Goyal, Bojanowski, and Joulin]{caron2020unsupervised}
Mathilde Caron, Ishan Misra, Julien Mairal, Priya Goyal, Piotr Bojanowski, and Armand Joulin.
\newblock Unsupervised learning of visual features by contrasting cluster assignments.
\newblock In \emph{NeurIPS}, 2020.

\bibitem[Caron et~al.(2021)Caron, Touvron, Misra, J\'egou, Mairal, Bojanowski, and Joulin]{caron2021emerging}
Mathilde Caron, Hugo Touvron, Ishan Misra, Herv\'e J\'egou, Julien Mairal, Piotr Bojanowski, and Armand Joulin.
\newblock Emerging properties in self-supervised vision transformers.
\newblock In \emph{ICCV}, 2021.

\bibitem[Changpinyo et~al.(2021)Changpinyo, Sharma, Ding, and Soricut]{cc12m}
Soravit Changpinyo, Piyush Sharma, Nan Ding, and Radu Soricut.
\newblock Conceptual 12m: Pushing web-scale image-text pre-training to recognize long-tail visual concepts.
\newblock In \emph{CVPR}, 2021.

\bibitem[Chen et~al.(2021)Chen, Hu, Wu, Jiang, and Wang]{Chen21learning}
Jiacheng Chen, Hexiang Hu, Hao Wu, Yuning Jiang, and Changhu Wang.
\newblock Learning the best pooling strategy for visual semantic embedding.
\newblock In \emph{CVPR}, 2021.

\bibitem[Chen et~al.(2013)Chen, Shrivastava, and Gupta]{Chen2013NEIL}
Xinlei Chen, Abhinav Shrivastava, and Abhinav Gupta.
\newblock Neil: Extracting visual knowledge from web data.
\newblock In \emph{ICCV}, 2013.

\bibitem[Chen et~al.(2014)Chen, Shrivastava, and Gupta]{Chen2014enrich}
Xinlei Chen, A. Shrivastava, and A. Gupta.
\newblock Enriching visual knowledge bases via object discovery and segmentation.
\newblock In \emph{CVPR}, 2014.

\bibitem[Choe et~al.(2020)Choe, Oh, Lee, Chun, Akata, and Shim]{choe2020cvpr}
Junsuk Choe, Seong~Joon Oh, Seungho Lee, Sanghyuk Chun, Zeynep Akata, and Hyunjung Shim.
\newblock Evaluating weakly supervised object localization methods right.
\newblock In \emph{CVPR}, 2020.

\bibitem[Desai and Johnson(2021)]{Desai21virtex}
Karan Desai and Justin Johnson.
\newblock Virtex: Learning visual representations from textual annotations.
\newblock In \emph{CVPR}, 2021.

\bibitem[Deselaers et~al.(2010)Deselaers, Alexe, and Ferrari]{Deselaers2010eccv}
Thomas Deselaers, Bogdan Alexe, and Vittorio Ferrari.
\newblock Localizing objects while learning their appearance.
\newblock In \emph{ECCV}, 2010.

\bibitem[Dosovitskiy et~al.(2021)Dosovitskiy, Beyer, Kolesnikov, Weissenborn, Zhai, Unterthiner, Dehghani, Minderer, Heigold, Gelly, Uszkoreit, and Houlsby]{Dosovitskiy2021}
Alexey Dosovitskiy, Lucas Beyer, Alexander Kolesnikov, Dirk Weissenborn, Xiaohua Zhai, Thomas Unterthiner, Mostafa Dehghani, Matthias Minderer, Georg Heigold, Sylvain Gelly, Jakob Uszkoreit, and Neil Houlsby.
\newblock An image is worth 16x16 words: Transformers for image recognition at scale.
\newblock In \emph{ICLR}, 2021.

\bibitem[Faghri et~al.(2018)Faghri, Fleet, Kiros, and Fidler]{Faghri2018Faghri}
Fartash Faghri, David~J. Fleet, Jamie~Ryan Kiros, and Sanja Fidler.
\newblock Vse++: Improving visual-semantic embeddings with hard negatives.
\newblock In \emph{BMVC}, 2018.

\bibitem[Fellbaum(1998)]{wordnet}
Christiane Fellbaum.
\newblock \emph{WordNet: An Electronic Lexical Database}.
\newblock Bradford Books, 1998.

\bibitem[Fini et~al.(2021)Fini, Sangineto, Lathuilière, Zhong, Nabi, and Ricci]{Fini_2021_ICCV}
Enrico Fini, Enver Sangineto, Stéphane Lathuilière, Zhun Zhong, Moin Nabi, and Elisa Ricci.
\newblock A unified objective for novel class discovery.
\newblock In \emph{ICCV}, 2021.

\bibitem[Frome et~al.(2013)Frome, Corrado, Shlens, Bengio, Dean, Ranzato, , and Mikolov]{Frome2013devise}
Andrea Frome, Gregory~S. Corrado, Jonathon Shlens, Samy Bengio, Jeffrey Dean, Marc{\textquotesingle}Aurelio Ranzato, , and Tomas{\"u} Mikolov.
\newblock Devise: A deep visual-semantic embedding model.
\newblock In \emph{NeurIPS}, 2013.

\bibitem[Goyal et~al.(2017)Goyal, Khot, Summers-Stay, Batra, and Parikh]{Goyal2017Making}
Yash Goyal, Tejas Khot, Douglas Summers-Stay, Dhruv Batra, and Devi Parikh.
\newblock Making the v in vqa matter: Elevating the role of image understanding in visual question answering.
\newblock In \emph{CVPR}, 2017.

\bibitem[Han et~al.(2019)Han, Vedaldi, and Zisserman]{han2019learning}
Kai Han, Andrea Vedaldi, and Andrew Zisserman.
\newblock Learning to discover novel visual categories via deep transfer clustering.
\newblock In \emph{ICCV}, 2019.

\bibitem[Han et~al.(2020)Han, Rebuffi, Ehrhardt, Vedaldi, and Zisserman]{han20automatically}
Kai Han, Sylvestre-Alvise Rebuffi, Sebastien Ehrhardt, Andrea Vedaldi, and Andrew Zisserman.
\newblock Automatically discovering and learning new visual categories with ranking statistics.
\newblock In \emph{ICLR}, 2020.

\bibitem[Han et~al.(2021)Han, Rebuffi, Ehrhardt, Vedaldi, and Zisserman]{han21autonovel}
Kai Han, Sylvestre-Alvise Rebuffi, Sebastien Ehrhardt, Andrea Vedaldi, and Andrew Zisserman.
\newblock Autonovel: Automatically discovering and learning novel visual categories.
\newblock \emph{IEEE TPAMI}, 2021.

\bibitem[He et~al.(2016)He, Zhang, Ren, and Sun]{he2016deep}
Kaiming He, Xiangyu Zhang, Shaoqing Ren, and Jian Sun.
\newblock Deep residual learning for image recognition.
\newblock In \emph{CVPR}, 2016.

\bibitem[H\'enaff et~al.(2022)H\'enaff, Koppula, Shelhamer, Zoran, Jaegle, Zisserman, Carreira, and Arandjelovi\'c]{Henaff22}
O.~J. H\'enaff, S. Koppula, E. Shelhamer, D. Zoran, A. Jaegle, A. Zisserman, J. Carreira, and R Arandjelovi\'c.
\newblock Object discovery and representation networks.
\newblock In \emph{ECCV}, 2022.

\bibitem[Ji et~al.(2019)Ji, Henriques, and Vedaldi]{ji2019invariant}
Xu Ji, Jo{\~a}o~F Henriques, and Andrea Vedaldi.
\newblock Invariant information clustering for unsupervised image classification and segmentation.
\newblock In \emph{ICCV}, 2019.

\bibitem[Jia et~al.(2021{\natexlab{a}})Jia, Yang, Xia, Chen, Parekh, Pham, Le, Sung, Li, and Duerig]{Jia2021align}
Chao Jia, Yinfei Yang, Ye Xia, Yi-Ting Chen, Zarana Parekh, Hieu Pham, Quoc~V. Le, Yun-Hsuan Sung, Zhen Li, and Tom Duerig.
\newblock Scaling up visual and vision-language representation learning with noisy text supervision.
\newblock In \emph{ICML}, 2021{\natexlab{a}}.

\bibitem[Jia et~al.(2021{\natexlab{b}})Jia, Han, Zhu, and Green]{jia21joint}
Xuhui Jia, Kai Han, Yukun Zhu, and Bradley Green.
\newblock Joint representation learning and novel category discovery on single- and multi-modal data.
\newblock In \emph{ICCV}, 2021{\natexlab{b}}.

\bibitem[Karpathy and Fei-Fei(2015)]{Karpathy2015}
Andrej Karpathy and Li Fei-Fei.
\newblock Deep visual-semantic alignments for generating image descriptions.
\newblock In \emph{CVPR}, 2015.

\bibitem[Khosla et~al.(2011)Khosla, Jayadevaprakash, Yao, and Fei-Fei]{KhoslaYaoJayadevaprakashFeiFei_FGVC2011}
Aditya Khosla, Nityananda Jayadevaprakash, Bangpeng Yao, and Li Fei-Fei.
\newblock Novel dataset for fine-grained image categorization.
\newblock In \emph{CVPRW}, 2011.

\bibitem[Krizhevsky and Hinton(2009)]{Krizhevsky09cifar}
Alex Krizhevsky and Geoffrey Hinton.
\newblock Learning multiple layers of features from tiny images.
\newblock \emph{Technical report}, 2009.

\bibitem[Krizhevsky et~al.(2012)Krizhevsky, Sutskever, and Hinton]{NIPS2012AlexNet}
Alex Krizhevsky, Ilya Sutskever, and Geoffrey~E Hinton.
\newblock Imagenet classification with deep convolutional neural networks.
\newblock In \emph{NeurIPS}, 2012.

\bibitem[Liu et~al.(2024)Liu, Li, Wu, and Lee]{llava}
Haotian Liu, Chunyuan Li, Qingyang Wu, and Yong~Jae Lee.
\newblock Visual instruction tuning.
\newblock 2024.

\bibitem[Lloyd(1982)]{LLoyd_kmeans}
Stuart~P. Lloyd.
\newblock Least squares quantization in pcm.
\newblock \emph{IEEE Transactions on Information Theory}, 1982.

\bibitem[Lu et~al.(2019)Lu, Batra, Parikh, and Lee]{Lu2019ViLBERT}
Jiasen Lu, Dhruv Batra, Devi Parikh, and Stefan Lee.
\newblock Vilbert: Pretraining task-agnostic visiolinguistic representations for vision-and-language tasks.
\newblock In \emph{NeurIPS}, 2019.

\bibitem[MacQueen(1967)]{MackQueen67_Kmeans}
James MacQueen.
\newblock Some methods for classification and analysis of multivariate observations.
\newblock In \emph{Proceedings of the Fifth Berkeley Symposium on Mathematical Statistics and Probability}, 1967.

\bibitem[Ouldnoughi et~al.(2023)Ouldnoughi, Kuo, and Kira]{clipgcd}
Rabah Ouldnoughi, Chia-Wen Kuo, and Zsolt Kira.
\newblock Clip-gcd: Simple language guided generalized category discovery.
\newblock \emph{arXiv preprint arXiv:2305.10420}, 2023.

\bibitem[Radford et~al.(2021)Radford, Kim, Hallacy, Ramesh, Goh, Agarwal, Sastry, Askell, Mishkin, Clark, Krueger, and Sutskever]{Radford2021Learning}
Alec Radford, Jong~Wook Kim, Chris Hallacy, Aditya Ramesh, Gabriel Goh, Sandhini Agarwal, Girish Sastry, Amanda Askell, Pamela Mishkin, Jack Clark, Gretchen Krueger, and Ilya Sutskever.
\newblock Learning transferable visual models from natural language supervision.
\newblock In \emph{ICML}, 2021.

\bibitem[Ridnik et~al.(2021)Ridnik, Ben-Baruch, Noy, and Zelnik-Manor]{ridnik2021imagenet21k}
Tal Ridnik, Emanuel Ben-Baruch, Asaf Noy, and Lihi Zelnik-Manor.
\newblock Imagenet-21k pretraining for the masses.
\newblock \emph{ArXiv e-prints}, 2021.

\bibitem[Ronen et~al.(2022)Ronen, Finder, and Freifeld]{Ronen_DeepDPM}
Meitar Ronen, Shahaf~E. Finder, and Oren Freifeld.
\newblock Deepdpm: Deep clustering with an unknown number of clusters.
\newblock In \emph{CVPR}, 2022.

\bibitem[Russakovsky et~al.(2015)Russakovsky, Deng, Su, Krause, Satheesh, Ma, Huang, Karpathy, Khosla, Bernstein, Berg, and Fei-Fei]{ILSVRC15}
Olga Russakovsky, Jia Deng, Hao Su, Jonathan Krause, Sanjeev Satheesh, Sean Ma, Zhiheng Huang, Andrej Karpathy, Aditya Khosla, Michael Bernstein, Alexander~C. Berg, and Li Fei-Fei.
\newblock {ImageNet Large Scale Visual Recognition Challenge}.
\newblock \emph{IJCV}, 2015.

\bibitem[Simonyan and Zisserman(2015)]{Simonyan15vgg}
Karen Simonyan and Andrew Zisserman.
\newblock Very deep convolutional networks for large-scale image recognition.
\newblock In \emph{ICLR}, 2015.

\bibitem[Stephan et~al.(2024)Stephan, Miklautz, Sidak, Wahle, Gipp, Plant, and Roth]{text_guided_cluster}
Andreas Stephan, Lukas Miklautz, Kevin Sidak, Jan~Philip Wahle, Bela Gipp, Claudia Plant, and Benjamin Roth.
\newblock Text-guided image clustering.
\newblock In \emph{EACL}, 2024.

\bibitem[Van~Gansbeke et~al.(2020)Van~Gansbeke, Vandenhende, Georgoulis, Proesmans, and Van~Gool]{vangansbeke2020scan}
Wouter Van~Gansbeke, Simon Vandenhende, Stamatios Georgoulis, Marc Proesmans, and Luc Van~Gool.
\newblock Scan: Learning to classify images without labels.
\newblock In \emph{ECCV}, 2020.

\bibitem[Vaze et~al.(2022)Vaze, Han, Vedaldi, and Zisserman]{vaze2022gcd}
Sagar Vaze, Kai Han, Andrea Vedaldi, and Andrew Zisserman.
\newblock Generalized category discovery.
\newblock In \emph{CVPR}, 2022.

\bibitem[Vo et~al.(2019)Vo, Bach, Cho, Han, LeCun, P\'{e}rez, and Ponce]{vo19unsup}
Huy~V. Vo, Francis Bach, Minsu Cho, Kai Han, Yann LeCun, Patrick P\'{e}rez, and Jean Ponce.
\newblock Unsupervised image matching and object discovery as optimization.
\newblock In \emph{CVPR}, 2019.

\bibitem[Vo et~al.(2021)Vo, Sizikova, Schmid, P\'{e}rez, and Ponce]{NEURIPS2021_8bf1211f}
Van~Huy Vo, Elena Sizikova, Cordelia Schmid, Patrick P\'{e}rez, and Jean Ponce.
\newblock Large-scale unsupervised object discovery.
\newblock In \emph{NeurIPS}, 2021.

\bibitem[Wah et~al.(2011{\natexlab{a}})Wah, Branson, Welinder, Perona, and Belongie]{WahCUB_200_2011}
Catherine Wah, Steve Branson, Peter Welinder, Pietro Perona, and Serge Belongie.
\newblock {The Caltech-UCSD Birds-200-2011 Dataset}.
\newblock Technical Report CNS-TR-2011-001, California Institute of Technology, 2011{\natexlab{a}}.

\bibitem[Wah et~al.(2011{\natexlab{b}})Wah, Branson, Welinder, Perona, and Belongie]{cub200}
Catherine Wah, Steve Branson, Peter Welinder, Pietro Perona, and Serge Belongie.
\newblock The caltech-ucsd birds-200-2011 dataset.
\newblock \emph{Technical Report CNS-TR-2011-001, California Institute of Technology}, 2011{\natexlab{b}}.

\bibitem[Yang et~al.(2017)Yang, Fu, Sidiropoulos, and Hong]{yang_icml_17}
Bo Yang, Xiao Fu, Nicholas~D. Sidiropoulos, and Mingyi Hong.
\newblock Towards k-means-friendly spaces: Simultaneous deep learning and clustering.
\newblock In \emph{ICML}, 2017.

\bibitem[Yue et~al.(2024)Yue, Chen, Geiping, Li, Goldstein, and Lim]{objectnext}
Kaiyu Yue, Bor-Chun Chen, Jonas Geiping, Hengduo Li, Tom Goldstein, and Ser-Nam Lim.
\newblock Object recognition as next token prediction.
\newblock 2024.

\bibitem[Zellers et~al.(2019)Zellers, Bisk, Farhadi, and Choi]{Zellers2019From}
Rowan Zellers, Yonatan Bisk, Ali Farhadi, and Yejin Choi.
\newblock From recognition to cognition: Visual commonsense reasoning.
\newblock In \emph{CVPR}, 2019.

\bibitem[Zhai et~al.(2022)Zhai, Wang, Mustafa, Steiner, Keysers, Kolesnikov, and Beyer]{Zhai2022lit}
Xiaohua Zhai, Xiao Wang, Basil Mustafa, Andreas Steiner, Daniel Keysers, Alexander Kolesnikov, and Lucas Beyer.
\newblock Lit: Zero-shot transfer with locked-image text tuning.
\newblock In \emph{CVPR}, 2022.

\bibitem[Zhao and Han(2021)]{zhao21novel}
Bingchen Zhao and Kai Han.
\newblock Novel visual category discovery with dual ranking statistics and mutual knowledge distillation.
\newblock In \emph{NeurIPS}, 2021.

\end{thebibliography}
}

\clearpage
\appendix
\section{Appendix}

\subsection{Evaluation on the predicted semantic names}

One perspective of the Semantic Category Discovery task is narrowing an unconstrained dictionary of $N$ possible names to the optimal set of $K$ categories for a given dataset (see sec. 3).
As such, a further possible metric is the overlap between the set of discovered class names and the ground truth set of semantic labels.
We evaluate the intersection over union (IoU) between the two sets and report results in~\cref{tab:iou}. 
We see that the reported IoUs are reasonably correlated with the sACC performance, substantiating our assumption that \textit{discovering} the true set of class names is a key challenge in unconstrained semantic labelling. 

\begin{table}[htb]
\footnotesize
\centering
\caption{\textbf{Evaluation on the predicted semantic names.} We measure the IoU between the ground-truth names and the predicted names for each dataset under both unsupervised and partially supervised settings.}
\label{tab:iou}
\resizebox{0.9\linewidth}{!}{ 
\begin{tabular}{lcc}
\toprule
            &  Unsupervised & Partially supervised \\
\midrule
ImageNet-100 & 0.290 & 0.575\\
\midrule
ImageNet-1K & 0.273 & 0.517\\
\midrule
Standord Dogs & 0.589 & 0.752\\
\midrule
CUB & 0.343 & 0.481\\
\bottomrule
\end{tabular}
}
\end{table}

\subsection{Effects of different initial clustering algorithms}
\label{ap:diff_km}
In~\cref{tab:diff_km}, we investigate the effects of using different non-parametric clustering algorithms, including $k$-means, semi-supervised $k$-means (SS-KM), and our constrained semi-supervised $k$-means (CSS-KM), on both CLIP and GCD features under the partially supervised setting, where the latter two clustering algorithms appear to be more demanding. 
Our semantic naming method can consistently improve the ACC, regardless of the non-parametric clustering algorithms or the features used for initial clustering. Among these three choices of algorithms, our CSS-KM gives the best semantic naming performance. Interestingly, we observe that the overall initial ACC of CSS-KM is not as good as SS-KM, but using our CSS-KM to provide initial clustering gives better ACC after semantic naming, validating the effectiveness of the size constraint we introduced in CSS-KM for the semantic voting. 
As the GCD feature extractor is trained jointly using supervised contrastive learning on the unlabelled data and self-supervised contrastive learning on all data for better visual representation, the initial clustering results on GCD features are better. We obtain a better semantic naming result with GCD features. As a stronger training signal is used for the `Old' classes, the ACC of $k$-means, SS-KM and CSS-KM are very high. Note the semantic naming is based on the CLIP feature, which is not biased to the labelled data; thus, our semantic naming gives the best `All' and `New' classes and the gap between `Old' and `New' is much smaller than the initial clustering based on the GCD feature.

\begin{table}[!htb]
\footnotesize
\caption{\textbf{Effects of different initial clustering algorithms.} Experiments under the partially supervised setting with CLIP and GCD features.}
\centering
\resizebox{1.\linewidth}{!}{ 
\begin{tabular}{lcccccc}
\toprule
& \multicolumn{3}{c}{CLIP} & \multicolumn{3}{c}{GCD} 
\\
\cmidrule(rl){2-4}
\cmidrule(rl){5-7}
Classes       & All      & Old     & New      & All      & Old & New       \\\midrule
$k$-means 
&  62.3   &  57.2 & 64.8    
&  78.2    & 89.0 & 72.8  \\
$k$-means + Semantic Naming
&  71.3   &  64.6 & 74.7    
&  80.9    &85.1 & 78.9  \\
\midrule
SS-KM 
&  68.1   &  74.1    & 65.0   
&  73.2    & 87.8 & 65.8  \\
SS-KM + Semantic Naming
&  71.2   &  80.7    & 66.4  
&74.0   &  81.9    & 70.0\\
\midrule
CSS-KM 
&  65.6  & 74.2    & 61.3   
& 78.7   & \textbf{92.9}   & 71.5 \\
CSS-KM + Semantic Naming
& \textbf{79.9}   &  \textbf{83.7}   & \textbf{78.0} 
&\textbf{83.0}   & 84.9    & \textbf{82.1} \\
\bottomrule
\end{tabular}
}
\label{tab:diff_km}
\end{table}

\subsection{Cluster size comparison}
\begin{figure*}
    \centering
    \includegraphics[width=\textwidth]{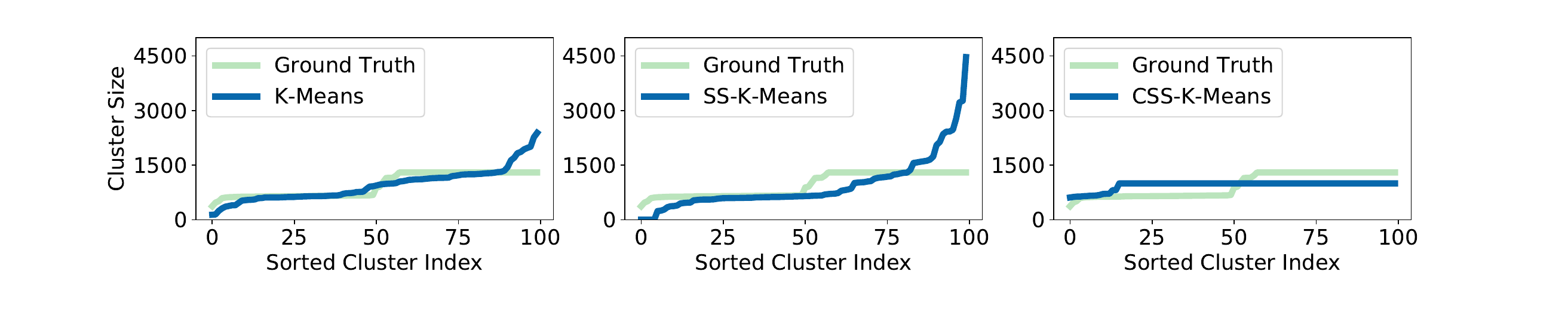}
    \caption{\textbf{Sorted cluster sizes obtained by different initial clustering algorithms.} Results are reported on ImageNet-100 dataset with DINO features.}
    \label{fig:scd_cluster_sizes}
\end{figure*}

Here, we compare the cluster sizes under the partially supervised setting with different clustering methods, in order to illustrate the benefits of our constrained algorithm (sec. 3.3.1). 
As can be seen, both $k$-means and SS-$k$-means can have clusters with very few instances (which is more severe in SS-$k$-means), making the voting from these few-instance clusters less reliable. 
In contrast, our improved CSS-$k$-means can give cluster sizes that are better aligned with the ground truth, thus benefiting the subsequent semantic voting process.
Note that in the partially supervised settings, rough cluster sizes can be estimated from the labelled data (and used to guide the CSS-$k$-means process).

\subsection{Qualitative analysis}

\begin{figure*}
    \centering
    \includegraphics[width=0.95\linewidth]{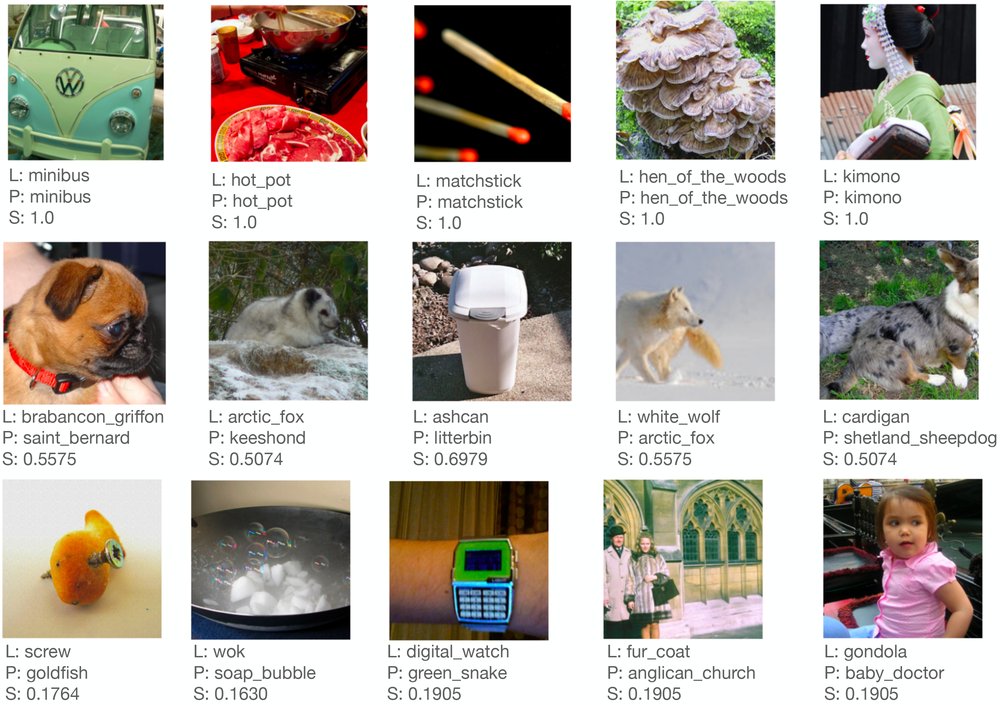}
    \caption{\textbf{Qualitative results on unlabelled instances from unknown classes.} Top row: correct predictions; Middle row: partially correct predictions; Bottom row: incorrect predictions. P: prediction; L: label; S: Soft semantic similarity score.}
    \label{fig:correct_partc_inc}
\end{figure*}

In~\cref{fig:correct_partc_inc}, we show qualitative results of our method on unlabelled images in  ImageNet-100. 
We report correct, partially correct, and incorrect cases for our method, where we use the soft semantic similarity score 
to bin examples.
For the partially correct predictions (middle row), we can see that the predictions are actually highly semantically relevant to the ground-truth names. The errors can be partially attributed to the pose (\eg, `brabancon griffon' vs `saint bernard'), occlusion (\eg, `cardigan'), and background clutter (\eg, `arctic fox') in the content. 
For the incorrect predictions (bottom row), we can see that the predictions are mostly semantically relevant to the content. 
Incorrect predictions can also be caused by spurious objects appearing in the image. 
For example, in the `screw' vs `goldfish' image, a goldfish-shaped object actually occupies a higher region of the image than the screw, leading to an incorrect prediction.
When multiple dominant objects appear in the images, the model may get confused on which one to predict. For example, in the `fur coat' vs `anglican church' image, both items appear in the image but the latter occupies a larger region. Similar for the `bondola' vs `baby doctor' image.
In these cases, multi-label evaluations of ImageNet may also be beneficial ~\cite{ridnik2021imagenet21k}.
Through these qualitative results, we can see that our method can produce reasonably good results which reflect the true semantics of the images, with failures generally occurring for understandable reasons. 

\begin{figure*}
    \centering
    \includegraphics[width=0.9\linewidth]{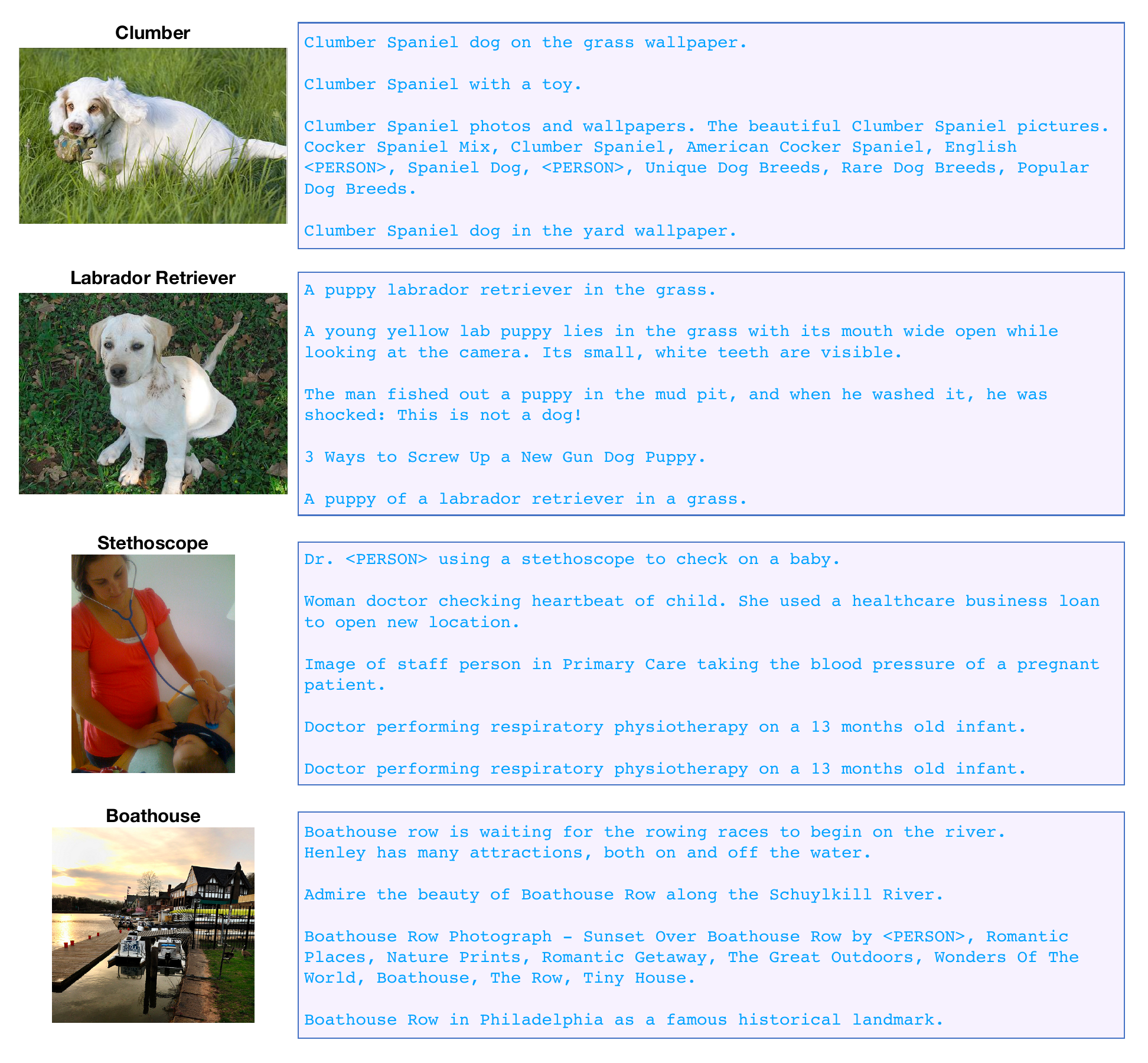}
    \caption{\textbf{Retrieved top-5 texts from CC12M dataset \cite{cc12m}.} The first two images are from \textit{Stanford Dogs} ~\cite{KhoslaYaoJayadevaprakashFeiFei_FGVC2011} and the last two images are from  \textit{ImageNet-1k}~\cite{ILSVRC15}. }
    \label{fig:text_example}
\end{figure*}

Additionally,  in \cref{fig:text_example},  we show examples of retrieved texts. We find that the retrieved text contains semantic information closely related to the objects in the images. 
For the first and second images, we can directly retrieve the corresponding dog breed, which can effectively improve the clustering accuracy. 
For the third and fourth images, we can extract supplementary information related to the target category, such as `Woman doctor checking heartbeat of child' (third image) and `Boathouse row is ready and waiting for the rowing races to begin on the river' (fourth image). 
This auxiliary information can help in interpreting the image content. 
Although we also retrieve some irrelevant information, such as `he was shocked: This is not a dog' for the fourth image, the proportion of such noise is relatively small and can be suppressed by other useful text content.

\subsection{Discussion on multi-modal LLMs}
In this paper, we present a method for semantic category discovery leveraging the pre-trained vision-language model, \ie, CLIP, in our experiments.
Our method is significantly different from multi-modal Large Language Models (LLMs) like LLaVA \cite{llava} and GPT4V \cite{gpt4}, which enjoy the massive scale pertaining and yield open-ended outputs that do not necessarily contain \textit{semantic names}. 
If one would like to obtain the semantic names based on such outputs, it might be possible to do so by applying certain post-processing techniques. 
In contrast, we only employ the pre-trained vision-language model with weak text-image pair supervision to align the visual and textual embedding and develop our framework specifically focusing on predicting semantic names. 
Moreover, a recent study \cite{objectnext} that employs multi-modal LLMs for object category recognition necessitates the model fine-tuning on extensive datasets, while our method is training-free, as it 
leverages unconstrained category vocabularies and visual-language semantic relationships for clustering, eliminating the need for any additional training procedures.

\subsection{Limitations and ethical considerations}
\label{sec:limitations}

In this work, we have tackled the task of assigning semantic names to images by automatically narrowing down an unconstrained vocabulary with a pre-trained vision language model.
Here, we highlight considerations when deploying such a method.

Firstly, we note that though our method outperforms existing baselines for this task, its absolute accuracy (both ACC and sACC) is quite low in absolute terms; for instance, we achieve roughly 30\% sACC (60\% Soft-sACC) on the unsupervised ImageNet setting.
This suggests that, even with internet-scale pre-training, current perception systems are not suitable for unconstrained deployment in the real-world. 
However, we hope that by highlighting this performance in the unconstrained setting, we can draw attention to an important research direction for the community. 

Moreover, we note that our method hinges on a model trained on internet-scale data (CLIP ~\cite{Radford2021Learning}) for which the training data is private and unavailable for inspection.
As such, we are unable to interrogate sources of possible bias in the model or else predict a-priori when its predictions may be unreliable. Hence, more transparent and controlled counterparts of CLIP models have demanded safety concerns in real-world deployment, such as AI-based surgical assistants. However, at this initial stage of exploring this problem. We believe adopting CLIP is a valid choice to verify the concepts of our methods for research purposes and other less safety concerned applications like entertainment. 

\subsection{License for datasets}
We carefully follow the licenses of the datasets in our experiments.
ImageNet~\cite{ILSVRC15} and Stanford Dogs~\cite{KhoslaYaoJayadevaprakashFeiFei_FGVC2011} apply the same license for non-commercial research use. CUB~\cite{cub200} dataset also allows non-commercial research use. 

\end{document}